\documentclass[journal]{IEEEtran}
\IEEEoverridecommandlockouts
\usepackage{cite}
\usepackage{amsmath,amssymb,amsfonts}
\usepackage{algorithm,float}
\usepackage{algorithmic}
\usepackage{graphicx}
\usepackage{textcomp}
\usepackage{xcolor}
\usepackage{subcaption}
\usepackage{stfloats}
\usepackage{bm,bbm}
\newtheorem{theorem}{Theorem}
\newtheorem{lemma}{Lemma}
\newtheorem{assumption}{Assumption}
\newtheorem{remark}{Remark}

\newtheorem{corollary}{Corollary}
\usepackage{url}
\usepackage{enumitem,kantlipsum}
\usepackage{color,soul}  
\graphicspath{{../figures/}}
\newcommand{\norm}[1]{\left\lVert#1\right\rVert} 
\def\BibTeX{{\rm B\kern-.05em{\sc i\kern-.025em b}\kern-.08em
    T\kern-.1667em\lower.7ex\hbox{E}\kern-.125emX}}
 
\DeclareMathOperator*{\argmax}{arg\,max}

\usepackage{scalerel,stackengine}
\stackMath
\newcommand\reallywidehat[1]{%
\savestack{\tmpbox}{\stretchto{%
  \scaleto{%
    \scalerel*[\widthof{\ensuremath{#1}}]{\kern-.6pt\bigwedge\kern-.6pt}%
    {\rule[-\textheight/2]{1ex}{\textheight}}
  }{\textheight}%
}{0.5ex}}%
\stackon[1pt]{#1}{\tmpbox}%
}
\begin{document}

\title{Federated Learning over Wireless {\color{black}IoT} Networks with Optimized Communication and Resources}

\author{
	Hao~Chen,~\IEEEmembership{Student~Member,~IEEE,}
	Shaocheng~Huang,
	Deyou~Zhang,
	Ming~Xiao,~\IEEEmembership{Senior~Member,~IEEE,}
	Mikael~Skoglund,~\IEEEmembership{Fellow,~IEEE, }
	and~H.~Vincent~Poor,~\IEEEmembership{Life Fellow,~IEEE }
 	\thanks{Hao Chen, Shaocheng Huang, Deyou Zhang, Ming Xiao and Mikael Skoglund are with the School of Electrical Engineering and Computer Science, Royal Institute of Technology (KTH), 10044 Stockholm, Sweden (email: \{haoch, shahua, deyou, mingx, skoglund\}@kth.se).
	
	H. Vincent Poor is with Department of Electrical and Computer Engineering, Princeton University, 08544 Princeton, USA (email: poor@princeton.edu).	
 	}	
		
}
  
\maketitle
 
\begin{abstract}

To leverage massive distributed data and computation resources, machine learning in the network edge is considered to be a promising technique especially for large-scale model training.
Federated learning (FL), as a paradigm of collaborative learning techniques, has obtained increasing research attention with the benefits of communication efficiency and improved data privacy. 
Due to the lossy communication channels and limited communication resources (e.g., bandwidth and power), it is of interest to investigate fast responding and accurate FL schemes over wireless systems.
Hence, we investigate the problem of jointly optimized communication efficiency and resources for FL over wireless {\color{black} Internet of things (IoT)} networks.
To reduce complexity, we divide the overall optimization problem into two sub-problems, i.e., the client scheduling problem and the resource allocation problem.
To reduce the communication costs for FL in wireless {\color{black}IoT} networks, a new client scheduling policy is proposed by reusing stale local model parameters.
To maximize successful information exchange over networks, 
a Lagrange multiplier method is first leveraged by decoupling variables including power variables, bandwidth variables and transmission indicators.
Then a linear-search based power and bandwidth allocation method is developed.
Given appropriate hyper-parameters, we show that the proposed communication-efficient federated learning (CEFL) framework converges at a strong linear rate.
Through extensive experiments, it is revealed that the proposed CEFL framework substantially boosts both the communication efficiency and learning performance of both training loss and test accuracy for FL over wireless {\color{black}IoT} networks compared to a basic FL approach with uniform resource allocation. 

\end{abstract}


\IEEEpeerreviewmaketitle

\section{Introduction}
\IEEEPARstart{W}{ith} the development of various emerging smart applications (e.g, augmented reality/virtual reality, autonomous driving and digital twin), the number of Internet of things (IoT) devices has increased explosively and the massive data generated from these connected IoT devices have led to a surging demand for very high communication rates in future wireless communications, such as the projected sixth generation (6G) mobile networks.
It is envisioned in \cite{iot_device} that by 2025, the active number of IoT devices is expected to be over 75 billion.
The massive amounts of data can bring diverse intelligent services due to the recent advances in artificial intelligence (AI) and large-scale machine learning (ML).
However, the data originating from massive IoT devices are commonly generated and stored in a distributed manner over wireless networks for a wide range of networked AI applications, e.g., smart grids\cite{smart_grid}, remote health monitoring \cite{remote_health}, etc.
Due to the nature of limited wireless communication resources as well as privacy concerns, it is often inefficient or impractical to directly collect all raw data of devices at a central entity (e.g., the cloud).
Alternatively, it is increasingly attractive to process data directly in edge clients for data analysis and inference by leveraging edge computing and intelligence with data kept locally.

In the regime of distributed machine learning, federated learning (FL), first coined by Google in 2016 \cite{fedavg},
is a paradigm of distributed ML, which pushes the computation of AI applications into edge clients.
Therefore, FL decouples the ability of ML from the need to reveal the data to a centralized location, which helps mitigate privacy and latency concerns.
During the training process of FL, in which edge clients seek to train a common ML model, each client periodically transmits its locally derived model parameters to a central parameter server (PS). A set of global model parameters are updated in the PS according to aggregation strategies such as federated averaging algorithm ($\textit{FedAvg}$) [3], and the PS then sends its updated global model parameters to clients for their local model updates. 
Compared with the traditional data-sharing based collaborative learning, both communication efficiency and user data privacy are significantly improved in FL.
Since the ML parameters are frequently exchanged between the PS and the edge clients over a wireless network, the performance of FL is largely constrained by the properties of wireless communication networks, which can be unstable and may even fluctuate significantly over time because of the limited wireless resources (e.g., bandwidth) and unreliable wireless channels.
Thus, this calls for a new design principle for FL
from both learning and wireless communication perspectives.

\subsection{Prior Work}
Since the proposal of FL \cite{nilsson2018performance, fedavg}, there has been an increasing number of studies related to the implementation of FL over wireless networks \cite{bonawitz2019towards, zhu2020toward, 9062302}.
Specifically, the authors in \cite{bonawitz2019towards} report on a system design of FL algorithms in the domain of Android mobile phones and sketch the challenges and corresponding solutions.

Despite the advantages of FL in terms of communication overheads and user data privacy over the traditional data-sharing based collaborative learning, the implementation of FL over wireless networks still suffers from bottlenecks.
More specifically, since multiple communication rounds are required to reach a desired ML accuracy, especially when the number of participating clients is comparably large, the communication costs incurred by unreliable wireless transmission become non-negligible in wireless FL systems.
To reduce communication overhead in distributed ML, various learning algorithms have been proposed in recent years \cite{prakash2020coded, codeadmm, quantization1, fl_quan, sparsfication, sparsfication2, chen2018lag, hosein, lin2017deep}.
Among these efforts, one research direction is to reduce the communication footprint in the uploading phase to make the model training communication efficient.
Typical approaches in this direction range from i) compressing the uploaded gradients via coding \cite{prakash2020coded, codeadmm}, quantization \cite{quantization1, fl_quan} and sparsification \cite{sparsfication, sparsfication2}, ii) limiting the model sharing by only updating clients with significant training improvement \cite{chen2018lag, hosein}, or iii) accelerating the training process by adopting a momentum method in the sparse update \cite{lin2017deep}.
More specifically, a lazily aggregated gradient approach is proposed in \cite{chen2018lag} to skip unnecessary uploads, among which communication censoring schemes are developed to avoid less informative local updates so as to reduce the communication burden.
It is also worth mentioning that the impact of network resources on the learning performance is not considered in any of those methods.

In addition to communication overhead, another series of work have focused on resource allocation in order to optimize the FL learning performance \cite{shi2020joint, zeng2020energy, nishio2019client, amiri2020update, yang2019scheduling, chen2020joint, chen2020convergence, xia2020multi, 8793221, amiri2021federated, 9187874, incentive_FL, chen2021communication}.
To improve wireless network efficiency, considerable research has been carried out and two main research directions play crucial roles including admission control and device scheduling \cite{nishio2019client, amiri2020update, yang2019scheduling, chen2020joint, chen2020convergence, xia2020multi, 8793221, amiri2021federated} and resource scheduling management (e.g., spectrum and power) \cite{chen2020joint, chen2020convergence}.
In \cite{nishio2019client}, a new FedCS protocol is proposed to schedule as many devices as possible in a limited time frame.
Another device scheduling policy is proposed in \cite{amiri2020update}, among which the channel conditions and the significance of the local model updates are jointly considered.
Nonetheless, these proposed policies are only evaluated via experiments and the convergence performance has not been theoretically analyzed.
To characterize the performance of FL in wireless networks, an analytical model with regard to the FL convergence rate has been developed and the impacts have been evaluated by three different client scheduling policies, i.e., random scheduling, round-robin, and proportional fair \cite{yang2019scheduling}.
By building a connection between the wireless resource allocation and the FL learning performance, the authors in \cite{chen2020joint, chen2020convergence} propose to optimize the user selection and power allocation to minimize the FL training loss. 
Despite of all these results, the existing methods often still involve high overhead both in computation and communication, especially for large-scale ML.

\subsection{Contributions}
Motivated by the above observations, we investigate FL with limited wireless resources. 
We study the problem of jointly optimizing resource and learning performance for reducing communication costs and improving learning performance in wireless FL systems.
Different from existing results, in what follows, we will study FL over wireless {\color{black}IoT networks} from the aspect of communication efficiency and wireless resource optimization co-design.
Particularly, a communication efficient federated learning (CEFL) scheme is proposed for wireless FL systems jointly taking communication efficiency and resource optimization into account.
The main contributions of our work can be summarized as follows:
\begin{itemize}
	
	\item We aim at communication efficient FL over wireless {\color{black}IoT} networks with limited resources.
    The joint optimization problem on communication efficiency and resource allocation is first formulated and then decoupled into a client scheduling sub-problem and a resource allocation sub-problem considering both bandwidth and power constraints.
	
	\item To reduce the communication costs of FL in wireless {\color{black}IoT} networks, a communication-efficient client scheduling policy is proposed by limiting communication exchanges and reusing stale local model parameters. 
	To optimize the resource allocation at each communication round of FL training, 
	the Lagrange multiplier method is leveraged to reformulate the resource optimization problem and an optimal solution based on linear search method is then derived.
	
	\item We investigate the convergence and communication properties of the proposed CEFL algorithm both analytically and by simulation. 
	Given a proper hyper-parameter, we show that CEFL achieves a strong linear convergence rate and $O\left( \log{ \frac{1}{\epsilon}}\right )$ communication loads, where $\epsilon$ is the target accuracy.
	In addition, the relation between the learning performance and wireless resources, namely bandwidth and power is theoretically analyzed.
	Experiment results also indicate that the proposed framework is communication efficient and resource optimized over wireless {\color{black}IoT} networks. Our CEFL algorithm outperforms the vanilla FL approach both in communication overhead and training and test performance.

\end{itemize}
The rest of this paper is structured as follows. 
Section \ref{system_model} describes the system model, while Section \ref{proposed_algorithm} discusses the design of our proposed FL algorithm optimized for the underlying wireless {\color{black}IoT} network.
In Section \ref{section:analysis}, we characterize the performance of our proposed framework over a wireless channel, which is validated via experiments in Section \ref{sec:results}.
We conclude the paper in Section \ref{conclusion} and technical proofs are provided in the Appendix.

\subsection{Notation}
We adopt the following notation in this paper. 
We denote expectation by $\mathbb{E}\left[\cdot\right]$.
$|\cdot|$ is the absolute value. 
$\norm{\cdot}$ denotes the $\boldsymbol{\ell}_2$-norm of a vector. 
$| \mathcal{N}_e^t |$ represents the cardinality of set $\mathcal{N}_e^t$. $\lceil{\cdot} \rceil$ is the ceiling function. 
$\nabla f(\cdot)$ denotes the gradient of a function $f$.
In addition, $\langle \cdot, \cdot \rangle $ denotes the inner product in a finite dimensional Euclidean space. 
$\bm{w}^*\in \mathcal{X}$ denotes the optimal solution to (\ref{eq:fl_ori}), where $\mathcal{X}$ is the domain. 
In addition, we define $ G_{\mathcal{X}} \overset{\Delta}{=} \mathop{ \text{sup}}\nolimits_{\bm{w}_x, \bm{w}_y \in \mathcal{X}} \norm{\bm{w}_x - \bm{w}_y}$.

\section{System Model and Problem Formulation}\label{system_model} 
\begin{figure} [b] 
	\vskip 0.2in
	\begin{center}
		\centerline{\includegraphics[width=88mm]{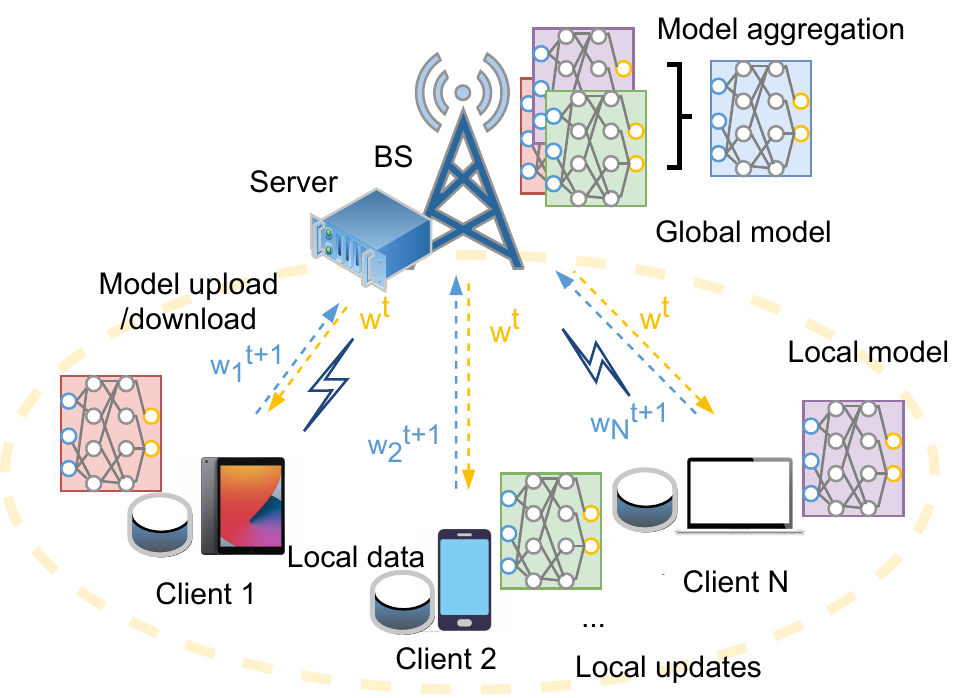}}
		\caption{An example of FL over the {\color{black} wireless IoT} network with multiple clients and a BS.}
	    \label{fig_network}
	\end{center}
	\vskip -0.2in
\end{figure}
In this section, we describe the framework of FL over wireless multi-client systems. 
We will discuss the network model, learning model, communication model and problem formulation.  
For ease of illustration, the notations used frequently in this paper are summarized in Table \ref{tab:table_parameters}.
\begin{table*}[t]
    \caption{Summary of Main Notation (at communication round $t$)}
    \label{tab:table_parameters}
    \centering
    \fontsize{9}{8}\selectfont
    \begin{tabular}{|c|c|c|c|}
        \hline 
        Notation & Description & Notation & Description \\
        \hline
        \hline
        $\mathcal{N}$ & Set of clients, $\mathcal{N}=\{1,...,N\}$ & $B^t$ & Total available bandwidth resource\\ \hline
        $\mathcal{N}_e^t$ & Set of scheduled clients & $\bm{B}^t$ & Bandwidth allocation vector\\ \hline
        $\mathcal{N}_c^t$ & Set of in-active clients& $\bm{P}^t$ & Power allocation vector\\ \hline
        $\mathcal{D}_i$ & Set of private data in client $i$&$P_i^{\text{max}}$ & Maximum power limit of client $i$ \\ \hline
        $D_i$ & Number of training samples in client $i$  & $P_i^{\text{min}}$ & Minimum power limit of client $i$   \\ \hline
        $D$ & Total number of training data samples,$ ~\sum_{i \in \mathcal{N}} D_i =D$ &$c_i^t$ & Transmission rate of client $i$ \\ \hline
        $\bm{w}^t$ & Global FL model & $\bm{a}^t$ & Transmission indicating vector \\ \hline
        $\reallywidehat{\bm{w}}_i^{t}$ &Local copy of the local FL model of client $i$ at PS & $\Gamma^t$ & Time threshold \\ \hline
        $\widetilde{\bm{w}}_i^t$ & Local copy of the global FL model at client $i$ &$S$ &Transmitted packet size \\ \hline
        $\bm{\xi}_{i,l}$ &  Training sample $l$ of client $i$, $\mathcal{D}_i=\{ \bm{\xi}_{i,1},...,\bm{\xi}_{i,D_i} \}$ &$\tau_{i}^t$ & Communication time of client $i$   \\ \hline
    \end{tabular}
\end{table*}

\subsection{Network Model}

As depicted in Fig. \ref{fig_network}, we consider a general one-hop {\color{black}FL-supported wireless IoT} network with a base station (BS) and $N$ distributed clients denoted as the set $\mathcal{N}=\{1,.., N \}$.
In this system, the BS directly connects to the PS, which is equipped with computational resources to provide communication and computation services to the clients.
The clients represent {\color{black}IoT} sensors gathering data for an FL task, such as mobile devices or organizations, which are communicated with the BS via wireless links.
We assume that each client $i$ collects measurement data and owns a fraction of labeled training samples, which is denoted as $\mathcal{D}_i=\{\bm{\xi}_{i,l}\}_{l=1}^{D_i}$ with $D_i = |\mathcal{D}_i|$ data samples and $\bm{\xi}_{i,l}$ representing the $l$-th training sample at client $i, \forall i \in \mathcal{N}$.
The whole dataset is thus denoted by $\mathcal{D} = \bigcup_{i\in \mathcal{N} } \mathcal{D}_i $ with the total number of data samples $D = \sum_{i=1}^{N} D_i$.
We consider training an ML model of interest over this network (e.g., a classifier), where the PS and clients collaboratively build a shared model parameter for data analysis and inference by exchanging model parameters information while keeping all the data locally.

\subsection{Federated Learning Process}
In the FL system, a global learning model of interest is trained in a distributed manner among geographically dispersed clients and then aggregated in a central server (i.e., PS).
The goal of the training process is to find a model parameter $\bm{w} \in \mathbbm{R}^d$ with the objective of minimizing a loss function $f(\bm{w})$ on the whole dataset $\mathcal{D}$.
The global learning objective of the network can be expressed as
\begin{align} \label{eq:fl_ori}
\min_{\bm{w}} \left \{f(\bm{w})\right \} &\overset{\Delta}{=} \min_{\bm{w}} \frac{1}{D} \sum _{i=1}^{N} \sum_{l=1}^{D_i}F_i(\bm{w},\bm{\xi}_{i,l})\notag\\
& = \min_{\bm{w}} \sum_{i= 1}^{N} \frac{D_i}{D} f_i(\bm{w}),
\end{align}
where the local loss function $f_i(\bm{w})$ of client $i$ is defined as $f_i(\bm{w})\overset{\Delta}{=} \frac{1}{D_i}\sum _{l=1}^{D_i} F_i(\bm{w},\bm{\xi}_{i,l} )$ and $F_i(\bm{w},\bm{\xi}_{i,l} )$ characterizes the loss of the model parameter $\bm{w}$ on the training sample $\bm{\xi}_{i,l} $.

Our analysis is based on the widely used federated averaging ($\textit{FedAvg}$) algorithm \cite{fedavg}. 
The whole training process is periodical with an arbitrary number of communication rounds (denoted as $T$), each of which has $E$ local epochs.
Then the $t$-th communication round is described by the following phases:
\begin{enumerate}
    \item \textbf{Broadcasting phase:} The PS (located in BS) wirelessly broadcasts the global model parameter $\bm{w}^t$ to all clients in the $t$-th round;
    \item \textbf{Local updating phase:} After receiving the global model parameter, each client $i \in \mathcal{N}$ trains its local model $\bm{w}_i^{t+1}$ by applying $E$ epochs of the gradient descent (GD) method, i.e., 
    \begin{align} \label{eq:gd_method}
        \bm{w}_i^{t+1}=\bm{w}^t - \eta \nabla f_i(\bm{w}^t),
    \end{align}
    where $\eta$ is the learning rate and $\nabla f_i(\bm{w}^t)$ is the gradient of local loss function.
    Then client $i$ uploads its updated local parameter $\bm{w}_i^{t+1}$ back to the PS. 
    It is noted that alternative methods, such as stochastic gradient descent (SGD), can also be used for local updates;
    \item \textbf{Aggregating and averaging phase:} For general FL framework, once receiving all the local model parameters, the PS aggregates them and obtains an updated global model by
    \begin{align} 
    \bm{w}^{t+1} = \sum _{i=1}^{N} \frac{D_i}{D} \bm{w}_i^{t+1}.
    \end{align}
    \end{enumerate}
The FL learning process implies that the FL model parameters are iteratively exchanged between the edge clients and the PS over wireless networks. 

\subsection{Communication Model}

We consider FL over a wireless medium with limited bandwidth and power.
After local training, clients upload their local FL models to the BS via frequency-division multiple access (FDMA).
Therefore, the achievable rate of client $i$ at the $t$-th communication round is given by
\begin{equation}
    c_i^t=B_i^t \log_2\left(1+ \frac{P_i^t |h_i^t|^2}{B_i^t N_0}\right),
\end{equation}
where $B_i^t $ and $P_i^t$ are the allocated bandwidth and transmission power of client $i$, respectively. $|h_i^t|^2$ denotes the corresponding single-carrier block-fading channel gain, and $N_0$ denotes the noise power spectral density.
For simplicity, it is also assumed that for client $i$, $\bm{w}_i^{t+1}$ is transmitted as a single packet in the uplink.
Denoting by $S$ the packet size of transmitted FL model as \cite{chen2020joint},
the communication time from client $i$ to the BS can be then given by
\begin{equation} \label{eq:S_ci}
        \tau_{i}^t = \frac{S}{c_i^t}.
\end{equation}
Since the transmit power of the BS can be generally much higher than that of the client and the whole downlink bandwidth can be utilized to broadcast the global model $\bm{w}^t$, the latency of downlink transmission is ignored to simplify illustration \cite{infocom_NNguyen, 9261995, energy_fl_chen}.
Moreover, to capture the effect of random channel variations on the transmission of each local model parameter $\bm{w}_i^t$, 
we consider the current transmission failure if $\tau_i^t > \Gamma^t$ holds in a time duration $\Gamma^t $,
and the corresponding outage probability is defined as:
\begin{equation}
    p_i^t = Pr (\tau_i^t > \Gamma^t).
\end{equation}

\subsection{Problem Formulation}
To achieve fast learning, the FL training process typically schedules as many clients as possible at each communication round \cite{low_latency_FL}.
However, it is undesirable for all clients engaged in learning to transmit their fresh local FL models to the PS
especially when the updates are conveyed over a wireless medium with limited resources (e.g., transmit power and network bandwidth).
Having more clients scheduled and uploading local models simultaneously can result in large overheads in communication, more unstable connections, and higher latency,
which inevitably lead to learning task with less accuracy.
To this end, we aim for an optimal solution of joint client scheduling and their associated resource allocation scheme in each communication round to pursue the best learning performance.
By denoting the transmission indicating vector as $\bm{a}^t$, we formulate the following optimization problem with the objective of optimizing both communication and resource for FL over wireless {\color{black}IoT} networks:
\begin{subequations}\label{eq:opti1}
\begin{align}
 (\text{P-0} )&
  \max_{\bm{a}^t, \bm{B}^t, \bm{P}^t}  {\sum_{i\in \mathcal{N}} a_i^t} \tag{\ref{eq:opti1}} \label{eq:op_objective}\\
\textrm{s.t.~~~} 
&P_i^{\text{min}} \leq P_i^t \leq P_i^{\text{max}}, \forall i \in \mathcal{N},\label{eq:op_const10}\\
& \sum_{i=1}^{N} B_i^t \leq B^t, \forall i \in \mathcal{N},\label{eq:op_const20}\\
&a_i^t=\left\{\begin{aligned}
    &1, \tau_{i}^t \leq \Gamma^t,\\
    &0, \tau_{i}^t > \Gamma^t, \forall i \in \mathcal{N},
    \end{aligned} \right . \label{eq:op_const50}
\end{align}
\end{subequations} 
where the objective of (P-0) is to maximize the utilization of transmitted FL parameters (i.e., the number of successful transmission) while sustaining the learning performance.
Constraints (\ref{eq:op_const10}) and (\ref{eq:op_const20}) are the feasibility conditions on the power allocation of clients and the bandwidth limits, respectively.
Constraint (\ref{eq:op_const50}) represents the successful transmission condition.
Here $a_i^t = 1 $ represents the successful transmission of the fresh local model $\bm{w}_i^{t+1}$ from client $i$; otherwise, we have $a_i^t =0 $.

\section{Federated Learning Algorithm Design over Wireless {\color{black}IoT} Networks} \label{proposed_algorithm}
(P-0) is a non-convex optimization problem due to the nonconvexity of its objective function and constraint (\ref{eq:op_const50}). 
To solve problem (P-0), we decompose it into two sub-problems, i.e., i) determining the client scheduling policy at each communication round, and ii) deciding the optimal resource allocation scheme for the clients that have been selected from sub-problem i).
We refer to the first subproblem as the \textit{client scheduling} problem and the second subproblem as the \textit{resource allocation} problem.

\subsection{Client Scheduling}

With the rapid development of integrated circuits, local computation time can be several orders of magnitude shorter than communication time between the clients and PS\cite{ic}.
The vanilla FL framework can lead to large communication overheads (e.g., communication time) and it can be inefficient to sequentially update the trained models from all clients before global aggregating and averaging \cite{client_sel}.
Accordingly, a subgroup of clients can be actively selected to transmit their local FL models simultaneously.
Then the communication efficiency can be improved and the communication latency can be reduced.
To this end, client scheduling policy plays a crucial role in FL process especially 
when wireless resources are limited with.
With the goal of reducing communication overheads per communication round, a communication-efficient client selection policy will be developed below.

For the $\textit{FedAvg}$ method in (\ref{eq:fl_ori}), in communication round $t$, after receiving the global model parameters $\bm{w}^t$ from the PS, every client $i \in \mathcal{N}$ updates its local parameter $\bm{w}_i^{t+1}$ via (\ref{eq:gd_method}) for $E$ epochs and is activated to feed updated $\bm{w}_i^{t+1}$ back to the PS.
Instead of requesting fresh local model parameters from all clients in (\ref{eq:gd_method}), our client scheduling policy runs as follows.
During each communication round $t$, client with informative messages (i.e., $\bm{w}_i^t$) is enabled to upload its current new model parameters if the following selection criterion meets:
\begin{align} \label{eq:comm_cond}
    N^2 \eta^2 \norm{\nabla f_i(\bm{w}^t) - \nabla f_i(\widetilde{\bm{w}}_i^{t})}^2
    \geq \sum _{k=1}^K \delta_k \norm{\bm{w}^{t+1-k}- \bm{w}^{t-k}}^2,
\end{align}
where $\{ \delta_k \}_{k=1}^K$ and $K$ are pre-defined constants, $\nabla f_i(\bm{w}^t) - \nabla f_i(\widetilde{\bm{w}}_i^{t})$ is the gradient difference between two evaluations of $\nabla f_i(\bm{w})$ at current model parameter $\bm{w}^t$ and the previous round model parameter $\widetilde{\bm{w}}_i^{t}$.
This condition compares the new local gradient to the stale copy at the client:
Only when the gradient difference is larger than the recent changes in $\bm{w}$, the new local model will be transmitted. Otherwise, the PS will reuse the stale copy at the PS.
In addition, to avoid clients inactive for a long time, we force it to upload its local model parameters $\bm{w}_i^t$ to the PS if any client $i$ has not been active for transmitting fresh model parameters during the past $T_0$ communication rounds.
To this regard, we set a clock $T_i, i \in \mathcal{N}$ for each client $i$, counting the number of inactive communication rounds since last time it uploaded its local models.
Thus, it always holds that
\begin{align} \label{eq:time_cond}
    T_i \leq T_0, \forall i \in  \mathcal{N}.
\end{align}
Once the fresh local model $\bm{w}_i^{t+1}$ in client $i$ satisfies the above conditions (\ref{eq:comm_cond}) and (\ref{eq:time_cond}), it will be uploaded to the PS, whilst the PS in BS will reuse the outdated local model parameters from the rest of clients.
{
\color{black}
We will prove in the next section that the proposed client scheduling policy based algorithm can still converge in a linear rate and is communication efficient.
}
Then, on the PS, the current copy of $\bm{w}_i$ from client $i$, denoted by $\reallywidehat{\bm{w}}_i^{t+1}$, is updated as
\begin{align} \label{eq:local_cpy}
    &\reallywidehat{\bm{w}}_i^{t+1}:=\left\{\begin{aligned}
        &\bm{w}_i^{t+1}, ~a_i^t = 1;\\
    &\reallywidehat{\bm{w}}_i^{t}~~~,~a_i^t=0,
    \end{aligned}
    \right.
\end{align}
where $\reallywidehat{\bm{w}}_i^{t}$ is the local model of client $i$ from previous rounds.
Here $a_i^t = 1 $ implies that the server receives the fresh local model $\bm{w}_i^{t+1}$ from client $i$, otherwise, we have $a_i^t =0 $.
\subsection{Power and Bandwidth Allocation}

Once \textit{client scheduling} is determined, the remaining subproblem is the bandwidth and power allocation among these scheduled clients. 
Given the set of scheduled clients $\mathcal{N}_e^t$ in the $t$-th communication round, \textit{resource allocation} subproblem can be formulated as follows:

\begin{subequations}\label{eq:opti1_1}
\begin{align}
 (\text{P-1} )&
  \max_{\bm{a}^t, \bm{B}^t, \bm{P}^t }  {\sum_{i\in \mathcal{N}_e^t} a_i^t} \tag{\ref{eq:opti1_1}} \label{eq:op_objective}\\
\textrm{s.t.~~~} 
&P_i^{\text{min}} \leq P_i^t \leq P_i^{\text{max}}, \forall i \in \mathcal{N}_e^t,\label{eq:op_const1}\\
& \sum_{i=1}^{N} B_i^t \leq B^t, \forall i \in \mathcal{N}_e^t,\label{eq:op_const2}\\
&a_i^t=\left\{\begin{aligned}
    &1, \tau_{i}^t \leq \Gamma^t,\\
    &0, \tau_{i}^t > \Gamma^t, \forall i \in \mathcal{N}_e^t.
    \end{aligned} \right . \label{eq:op_const5}
\end{align}
\end{subequations} 
Problem (P-1) is a mixed integer non-linear programming (MINLP) problem due to the binary variable $\{a_i^t\}$ and continuous variables $\{B_i^t\}$ and $\{P_i^t\}$.
By introducing big-$M$ constant for constraint (\ref{eq:op_const5}), problem (P-1) can be equivalently rewritten as
\begin{subequations}\label{eq:opti2}
\begin{align}
 (\text{P-2} )&
  \min_{a_i^t \in \{0,1\}, \bm{B}^t, \bm{P}^t}  {-\sum_{i\in \mathcal{N}_e^t} a_i^t} \tag{\ref{eq:opti2}} \label{eq:op_objective2}\\
\textrm{s.t.~~~} 
&(\ref{eq:op_const1})- (\ref{eq:op_const2}),  \notag \\
& \tau_{i}^t \leq \Gamma^t + M \left( 1 - a_i^t \right),\label{eq:op_const22}\\
& \tau_{i}^t \geq  \Gamma^t - M a_i^t. \label{eq:op_const23}
\end{align}
\end{subequations} 
By relaxing each binary variable $a_i^t \in \{ 0, 1\}$ to a continuous variable  $\widetilde{a}_i^{t} \in [0,1], \forall i \in \mathcal{N}_e^t$, we simplify (P-2) to a non-linear programming problem, given by
\begin{subequations}\label{eq:opti3}
\begin{align}
 (\text{P-3} )&
  \min_{\widetilde{a}_i^{t} \in [0,1], \bm{B}^t, \bm{P}^t}  {-\sum_{i\in \mathcal{N}_e^t}  \widetilde{a}_i^{t}} \tag{\ref{eq:opti3}} \label{eq:op_objective3}\\
\textrm{s.t.~~~} 
&(\ref{eq:op_const1})- (\ref{eq:op_const2}),  \notag \\
& \tau_{i}^t \leq \Gamma^t + M \left( 1 - \widetilde{a}_i^{t} \right),\label{eq:op_const32}\\
& \tau_{i}^t \geq  \Gamma^t - M \widetilde{a}_i^{t}. \label{eq:op_const33}
\end{align}
\end{subequations} 
(P-3) is still non-convex, and we resort to the Karush-Kuhn-Tucker (KKT) conditions for building the relation between bandwidth and power allocation.
More specifically, we first construct the associated Lagrangian function of (P-3) as follows:
\begin{align}\label{eq:p3_La}
&\mathcal{L}_{\rho}(\bm{\widetilde{a}}^{t},\bm{B}^t, \bm{P}^t)\notag\\
&= -\sum_{i\in \mathcal{N}_e^t}  \widetilde{a}_i^{t} + \sum_{i\in \mathcal{N}_e^t} x_i \left ( \widetilde{a}_i^{t} -1 \right) +\sum_{i\in \mathcal{N}_e^t}  y_i \left (-  \widetilde{a}_i^{t} \right) \notag\\ 
&\quad+ \mu  \left ( \sum_{i\in \mathcal{N}_e^t} B_i^t - B^t  \right) + \sum_{i\in \mathcal{N}_e^t} p_i \left ( P_i^{\text{min}} - P_i^t \right) \notag\\
&\quad+ \sum_{i\in \mathcal{N}_e^t} l_i \left ( P_i^t  - P_i^{\text{max}} \right)+ \sum _{i\in \mathcal{N}_e^t} \lambda_i \left( \frac{S}{c_i^t} - \Gamma^t + M\left( \widetilde{a}_i^{t} -  1 \right) \right)\notag\\
&\quad+ \sum _{i\in \mathcal{N}_e^t} \upsilon_i \left( \Gamma^t -  M\widetilde{a}_i^{t} -  \frac{S}{c_i^t} \right),
\end{align}
where $\{x_i, y_i, p_i, l_i, \lambda_i, \upsilon_i, \mu | i\in \mathcal{N}_e^t \}$ are nonnegative Lagrangian multipliers.
The KKT conditions for (P-3) are written as
\begin{align}
    &\frac{\partial \mathcal{L}}{\partial \widetilde{a}_i^{t}} = -1 + x_i - y_i + \left(\lambda_i -\upsilon_i\right) M = 0,\label{eq:kkt_con1}\\
    & 0 \leqslant \widetilde{a}_i \leqslant 1, \label{eq:a_range}\\
    & x_i \left ( \widetilde{a}_i -1 \right) = 0,\label{eq:kkt_con2}\\
    & y_i \left (-  \widetilde{a}_i \right) = 0,\label{eq:kkt_con3}\\
    &\lambda_i \left( \frac{S}{c_i} - \Gamma^t + M\left( \widetilde{a}_i -  1 \right) \right) =0,\label{eq:kkt_con4}\\
    & \upsilon_i \left( \Gamma^t -  M\widetilde{a}_i -  \frac{S}{c_i} \right) = 0,\label{eq:kkt_con5} \forall i \in \mathcal{N}_e^t,
\end{align}
where the solution pair of primal vectors and dual vectors is denoted as 
($\bm{\widetilde{a}^{*}}, \bm{B^{*}}, \bm{P^{*}}$) and ($\bm{x^*}, \bm{y^*}, \bm{p^*}, \bm{l^*}, \bm{\lambda^*}, \bm{\upsilon^*}, \mu^* $).
Then, according to (\ref{eq:kkt_con1})-(\ref{eq:kkt_con5}),
two lemmas can be obtained, as detailed below.

\begin{lemma}\label{lemma1_new}
Given the solution ($\bm{\widetilde{a}^{*}}, \bm{B^{*}}, \bm{P^{*}}$) of (P-3), the relation among transmission indicator, bandwidth and power allocation is given by
\begin{align} 
    \frac{S}{c_i^{*}} = \Gamma^t + M\left( 1- \widetilde{a}_i^{*} \right), \forall i \in \mathcal{N}_e^t, \label{eq:proposition}
\end{align}
where $c_i^{*}$ is defined by $c_i^{*}=B_i^{*} \log_2\left(1+ \frac{P_i^{*} |h_i^t|^2}{B_i^{*} N_0}\right)$.
\end{lemma}
\begin{IEEEproof}
To prove this lemma, we split $\{\widetilde{a}_i^{*}\}$ into three cases and we will show that (\ref{eq:proposition}) is valid in every case. Specifically,
\textbf{Case} 1) If $0<\widetilde{a}_i^{*}<1 $ holds, it is easy to obtain $x_i^* = y_i^* = 0$, and $\left(\lambda_i^* -\upsilon _i^*\right) M = 1$, based on (\ref{eq:kkt_con1})-(\ref{eq:kkt_con5}). Thus, $\lambda_i^*>0, \upsilon_i^*=0$, and $\frac{S}{c_i^{*}} = \Gamma^t + M\left( 1- \widetilde{a}_i^{*} \right)$ are achieved, $\forall i \in \mathcal{N}_e^t$; 
\textbf{Case} 2) If $a_i^*=0$ holds, we obtain $x_i^* = 0, \left(\lambda_i^* -\upsilon _i^*\right) M =  1 + y_i^*$, followed by $\lambda_i^*>0, \upsilon_i^* = 0$, and $\frac{S}{c_i^{*}} = \Gamma^t + M, \forall i \in \mathcal{N}_e^t$; 
\textbf{Case} 3) If $a_i^*=1$ holds, the following are implied as $y_i^*=0,~ \upsilon_i^*=0,~ x_i^* + M \lambda_i^*  = 1 ,~\frac{S}{c_i^*} \leq \Gamma^t$, and $\lambda_i^*\left( \frac{S}{c_i^*} - \Gamma^t \right)=0$.
In this case, though $\frac{S}{c_i^*} < \Gamma^t$ with $\lambda_i^*=0$ satisfies the KKT conditions, such a solution consumes more resources than condition of $\frac{S}{c_i^*} = \Gamma^t$ with $\lambda_i^*=\geq 0$.
Since we aim to achieve a resource optimized FL, we choose $\frac{S}{c_i^*} = \Gamma^t$, which proves (\ref{eq:proposition}) holds as well.
This completes the proof.
\end{IEEEproof}
\begin{remark}
(P-3) is a direct extension of the MINLP problem (P-1) or (P-2), and the proof of Lemma \ref{lemma1_new} shows that (\ref{eq:proposition}) also holds both in $\widetilde{a}_i^{*}=0$ and $\widetilde{a}_i^{*}=1$.
Thus, the relation among transmission indicator, allocated bandwidth and transmission power is also applicable for the initial problem (P-1) or (P-2), i.e., 
\begin{align} 
    \frac{S}{c_i^{*}} = \Gamma^t + M\left( 1- a_i^{*} \right), \forall i \in \mathcal{N}_e^t, \label{eq:remark1}
\end{align}
although it is an MINLP problem.
Particularly, for scheduled clients with successful transmission $a_i^{*}=1$, we have $S / c_i^{*}  =\Gamma^t .$
In addition, with the allocated power and transmission indicator, bandwidth allocation can be obtained directly via (\ref{eq:remark1}).
\end{remark}
\begin{algorithm}[t]
\caption{Linear Search Method for Resource Allocation} \label{algorithm:rsa_lsm}
\begin{algorithmic}[1]
    \STATE \textbf{Input}: $\mathcal{N}_e^t$, $B^t$,$\{ h_i^t, P_i^{\text{min}}, P_i^{\text{max}} | \forall i \in \mathcal{N}_e^t  \}$  ;
    \STATE \textbf{Initialize}: $U_{nc}^t = \left\{ i| i \in \mathcal{N}_e^t \right \}, U_{c}^t =   \emptyset$;
    \STATE construct the channel gain $|h_i^t|^2$ and sort it into a descending order as
        \begin{align} \label{eq:sort_order}
        |h_1^t|^2 \geq |h_2^t|^2\geq ... \geq|h_i^t|^2\geq...\geq |h_{|\mathcal{N}_e^t|}^t|^2, \forall i \in  \mathcal{N}_e^t;
    \end{align}
    \WHILE{$\sum_{i \in U_c^t}B_i^t \leq B^t$ and $U_{nc}^t \neq \emptyset$}
        \STATE find client $i$ whose $H$ is maximum in $U_{nc}^t$ as:
        \begin{align}
            &i = \argmax_{i \in U_c^t} \{ |h_i^t|^2 \};
        \end{align}
        \STATE set ${a}_i^{t} = 1$ and allocate power for client $i$ as $P_i^{t}=P_i^{\text{max}}$ and compute its required bandwidth $B_i^{t}$ via (\ref{eq:remark1});
        \STATE assign
            \begin{align}
                &U_{nc}^t = U_{nc}^t / \{i\}, U_{c}^t = U_{c}^t \cup \{ i \} ; \notag
            \end{align}
    \ENDWHILE
    \FOR{client $i \in U_{nc}^t$}
        \STATE set ${a}_i^{t} = 0$ and do not allocate network resources;
    \ENDFOR
\end{algorithmic} 
\end{algorithm}

\begin{lemma}\label{lemma2_new}
Given the selected client $i (i  \in \mathcal{N}_e^t) $ in communication round $t$,
its communication time $\tau_i^t$ is a decreasing function of $P_i^t$, $B_i^t$, and $|h_i^{t}|^2$.
\end{lemma}
\begin{IEEEproof}
The first derivative and second derivative of $c_i^t$ with respect to $P_i^t$ can be both proven to be larger than zero.
Thus, $c_i^t$ is an increasing and convex function of $P_i^t$. 
According to (\ref{eq:S_ci}), $\tau_i^t$ is then a decreasing function of the power $P_i^t$.
The same procedure works for $B_i^t$, and $|h_i^{t}|^2$, which proves Lemma \ref{lemma2_new}.
\end{IEEEproof}
Lemma \ref{lemma2_new} suggests that allocating larger transmission power $P_i^t$ contributes to less communication time $\tau_i^t$, which reduces the outage probability of transmission per communication round. 
In \textbf{Algorithm} \ref{algorithm:rsa_lsm}, we summarize our proposed resource allocation approach, i.e., linear search (LS) algorithm, which is exactly based on these two lemmas.
In step 3, we first sort the channel gains of the scheduled clients ($\forall i \in  \mathcal{N}_e^t$) in a descending order, which is denoted as $H \overset{\Delta}{=} |h_i^t|^2, \forall i \in  \mathcal{N}_e^t$.
Then, for client $i$ with the maximum channel gain in the un-allocated set $U_{nc}^t (i \in U_{nc}^t)$, we allocate its required power $P_i^t$ and bandwidth $B_i^t$ before categorizing it into the allocated set $U_{c}^t$ via steps 4-8.
The above steps are repeated until all scheduled clients are considered or all the available bandwidth and power resources are used up.
It is noted that according to the proposed LS method, those clients in the un-allocated set $U_{nc}^t$ will not be allocated bandwidth or transmission power resources.
Then, we give the performance analysis for the proposed LS method.
\begin{theorem}[Optimal solution] \label{theorem:1}
The proposed Algorithm 1 can provide an optimal solution for problem (P-1).
\end{theorem}
\begin{IEEEproof}
{\color{black}
The objective of problem (P-1) is to maximize the number of successful transmission.
Suppose that the solution based on Algorithm 1 can serve $K^*$ clients at most to successfully transmit their local FL models. 
For the convenience of explanation,  
based on (\ref{eq:sort_order}), we assume $K^*$ clients from $U_0^t=\{1,...,K^*\}$ are selected and the transmission indicating vector can be denoted as
$\bm{a}^*=\{\underbrace{1,...,1,...,1}_{K^*},\underbrace{0,...,0}_{N-K^*}\}$. 
With Lemmas \ref{lemma1_new} and \ref{lemma2_new}, we obtain
\begin{align} \label{eq: bw_opt}
    P_k^t:=\left\{\begin{aligned}
    &P_k^{\text{max}}, k \leq K^*,\\
    &P_i^{\text{min}},\text{otherwise};
    \end{aligned}
    \right.
    \sum _{k=1}^{K^*} B_k^t \leq B^t;
    \sum _{k=1}^{K^* + 1} B_k^t > B^t,
\end{align}
where $B_1^t\leq ...\leq B_{K^*}^t \leq B_{K^*+ 1}^t \leq ... \leq B_N^t, \forall k \in \mathcal{N}$.
Obviously, to support $K^*$ clients successfully transmit models, the resource allocation based on Algorithm 1 will consume the minimum bandwidth.
We assume that there exists another allocation scheme
in which $(K^* + 1)$ local fresh FL models can be successfully transmitted to PS.
Denote the active clients by $U_{1}^t (U_{1}^t \subset \mathcal{N}_e^t, |U_{1}^t|= K^*+1)$, then the following holds
\begin{align} \label{eq:scheme_u}
    P_l^{\text{min}} \leq P_l^t \leq P_l^{\text{max}}, \forall l \in U_{1}^t;
    \sum _{l \in U_{1}^t, |U_{1}^t|= K^*+1} B_l^t \leq B^t.
\end{align}
In (\ref{eq:scheme_u}), one possible solution can be
$\{P_l^{'} =P_l^{\text{max}}, B_l^{'} \leq B_l^t,\forall l \in U_{1}^t\}$ due to Lemma \ref{lemma2_new}, in which $\sum_{l \in U_{1}^t} B_l^{'} \leq \sum_{l \in U_{1}^t} B_l^t$ holds.
Then we could have
\begin{align}
    B^t \geq \sum_{l \in U_{1}^t} B_l^t \geq \sum_{l \in U_{1}^t} B_l^{'} \geq \sum_ {k=1}^{K^* + 1} B_k^t,
\end{align}
where it is shown to lead to a contradiction with (\ref{eq: bw_opt}).
Thus, the proposed Algorithm 1 can provide an optimal solution for (P-1), which proves Theorem 1. 
}
\end{IEEEproof}

Finally, we summarize the proposed CEFL framework in \textbf{Algorithm} \ref{algorithm:cefl} for clarity. 
For each communication round $t$, the PS broadcasts the global FL model $\bm{w}^t$ to all selected clients.  
Each client trains its local model after receiving $\bm{w}^t$ and independently decides whether or not to upload its own fresh local model via criteria (\ref{eq:comm_cond}) and (\ref{eq:time_cond}).
Upon receiving the updated models from the scheduled clients, the PS updates the global model with (\ref{eq:cefl_sum}).
The above steps (i.e., steps 4-19) are repeated until the stopping criterion is satisfied.
\begin{algorithm}[t]
\caption{ Communication-Efficient Federated Learning (CEFL) over Wireless {\color{black}IoT} Networks} \label{algorithm:cefl}
\begin{algorithmic}[1]
    \STATE \textbf{Input}: learning rate $\eta >0$, and constants $\{ \delta_k \}$;
	\STATE \textbf{Initialize}: $\{ \bm{w}^1, \reallywidehat{\bm{w}}_i^1, \widetilde{\bm{w}}_i^1, \nabla f_i(\widetilde{\bm{w}}_i^1),|\forall i \in \mathcal{N} \}$; 
    \FOR{$t=1,2,...$}
        \STATE Server broadcasts $\bm{w}^t$ to all clients;
        \STATE \ul{\textbf{Client Scheduling Policy:}}
        \FOR{each client $i\in \mathcal{N}$ \textbf{in parallel}}
            \STATE Receive $\bm{w}^t$ and compute $\nabla f_i(\bm{w}^t)$;
            \STATE Check condition (\ref{eq:comm_cond}) and (\ref{eq:time_cond});
            \IF{condition holds at client $i$}
                \STATE Update $\bm{w}_i^{t+1} $ based on (\ref{eq:gd_method}) and upload it;
                \STATE Update $\widetilde{\bm{w}}_i^{t+1} = \bm{w}^{t}$;
            \ELSE
                \STATE Update $\widetilde{\bm{w}}_i^{t+1} = \widetilde{\bm{w}}_i^{t}$ and upload nothing;
            \ENDIF
        \ENDFOR
        \STATE \textbf{update} the scheduling clients set $\mathcal{N}_e^t$;
        \STATE \ul{\textbf{Resource Allocation: call Algorithm 1;}}
        \STATE Server receives $\bm{w}_i^{t+1}$ and updates $\reallywidehat{\bm{w}}_i^{t+1}$ via (\ref{eq:local_cpy});
        \STATE Server compute $\bm{w}^{t+1}$ by
        \begin{equation} \label{eq:cefl_sum}
            \bm{w}^{t+1} := \sum _{i=1}^{N} \frac{D_i}{D} \reallywidehat{\bm{w}}_i^{t+1};
        \end{equation}
    	\STATE \textbf{until} the stopping criterion is satisfied.
    \ENDFOR
\end{algorithmic}  
\end{algorithm} 

\section{Convergence and Communication Analysis}\label{section:analysis}
In this section, we will first provide the theoretical analysis on the convergence of the proposed CEFL algorithm. 
Then we analyze the communication cost.
\subsection{Convergence Analysis}

To facilitate the convergence analysis, following \cite{infocom_NNguyen, 9261995, energy_fl_chen}, we assume that the BS-to-client transmission is error-free due to the rich power and bandwidth budget at the BS, and we evaluate the impact of noisy upload transmission on the convergence performance.
In addition, local epoch $E=1$ is considered for all clients to train a global FL model.
In the following, before analyzing the convergence of the CEFL algorithm, we first provide the following sufficient conditions, which are widely adopted in the analysis of decentralized optimization.



\begin{assumption}[Smoothness] \label{ass:smoothness}
The global loss function $f(\bm{x})$ is \textit{L-}smooth, i.e., for any $\bm{x}, \bm{y} \in\mathbbm{R}^{d}$,
\begin{equation}  \label{ass:smooth_eq1}
\begin{aligned}
\norm{ \nabla f(\bm{x}) -  \nabla f(\bm{y}) } \leqslant L \norm{  \bm{x} -\bm{y}  },
\end{aligned}
\end{equation}
and this is equivalent to 
\begin{equation} \label{ass:smooth_eq2}
f(\bm{x}) \leqslant f(\bm{y}) + \left\langle  \nabla f(\bm{y}), \bm{x}-\bm{y}  \right\rangle + \frac{L}{2} \norm{\bm{x}-\bm{y}}^2.
\end{equation}
\end{assumption}

\begin{assumption}[Coercivity] \label{ass:coverci}
The local loss function $f(\bm{x})$ is coercive over its feasible set $\mathcal{F}$, i.e., $f(\bm{x})\to \infty$ if $\bm{x}\in\mathcal{F}$ and $\|\bm{x} \|\to \infty$.  
The global loss function $f(\bm{x})$ is lower bounded over $\bm{x}\in\mathcal{F}$.
\end{assumption}


\begin{assumption}[Strong convexity]\label{assump:strong_convex}
The loss function $f_i(\bm{x})$ is $\mu$-$\textit{strongly}$ convex, satisfying that
\begin{equation}
    f(\bm{x}) \geq f(\bm{y}) + \left\langle  \nabla f(\bm{y}), \bm{x}-\bm{y}  \right\rangle + \frac{\mu}{2} \norm{\bm{x}-\bm{y}}^2,
\end{equation}
and 
\begin{equation} \label{eq:ass_grad_inequality}
    2\mu \left( f(\bm{x}) -f(\bm{x}^*) \right) \leq \norm{\nabla f(\bm{x})}^2.
\end{equation}
\end{assumption}
With these assumptions, we conclude the convergence properties of CEFL algorithm as follows.
\begin{lemma} \label{lemma_1}
  Suppose Assumptions 1 and 2 hold. Let $\{\bm{w}^t\}$ be the iterates generated by \textit{FedAvg} approach. If the learning rate satisfies $\eta = \frac{1}{L}$ and the outage probability is $p_i^t=0$, the \textit{FedAvg} update per communication round yields the following descent
  \begin{equation} \label{eq:lemma1_new}
      f(\bm{w}^{t+1}) \leq f(\bm{w}^t) - \Delta _{\textit{FedAvg}}^t,
  \end{equation}
  where $\Delta _{\textit{FedAvg}}^t \overset{\Delta}{=} \frac{1}{2L} \norm{\nabla f(\bm{w}^t)}^2$.
\end{lemma}
\begin{IEEEproof}
The proof of Lemma \ref{lemma_1} is similar to that in \cite{chen2018lag} and we omit it here due to space limitation.
\end{IEEEproof}
\begin{lemma} \label{lemma_2}
   Suppose Assumptions 1 and 2 hold. Let $\{\bm{w}^t\}$ be the iterates generated by 
    CEFL approach. If the learning rate satisfies $\eta = \frac{1}{L}$ and the outage probability is $p_i^t=0$, the CEFL update per communication round yields the following descent
   \begin{equation} \label{eq:lemma2}
       f(\bm{w}^{t+1}) \leq f(\bm{w}^t) - \Delta _{\textit{CEFL}}^t,
   \end{equation}
   where $\Delta _{\textit{CEFL}}^t$ is defined as
   \begin{align}
       &\Delta _{\textit{CEFL}}^t \overset{\Delta}{=} \notag\\
       &\frac{1}{2L} \norm{\nabla f(\bm{w}^t)}^2 - \frac{1}{2L} \norm{ \sum _{i \in \mathcal{N}_e^t} \frac{D_i}{D} \left [ \nabla f_i(\widetilde{\bm{w}}_i^t) - \nabla f_i(\bm{w}^t)\right] }^2.
   \end{align}
\end{lemma}
\begin{IEEEproof}
The proof of Lemma \ref{lemma_2} is similar to that in \cite{chen2018lag}, and we omit it here due to space limitation.
\end{IEEEproof}
\begin{remark} \label{remark_condition}
With the above lemmas, the rationale of (\ref{eq:comm_cond}) follows next.
Similar to the work in \cite{chen2018lag}, the proposed client scheduling policy selects the fresh local models by assessing its contribution to the loss function decrease.
To improve the communication efficiency of CEFL, each CEFL upload should bring more descent, i.e.,
\begin{align} \label{eq:descent_compare}
    \frac{\Delta _{\textit{CEFL}}^t}{| \mathcal{N}_e^t |} \geq \frac{\Delta _{\textit{FedAvg}}^t}{N}.
\end{align}
As stated in \cite{chen2018lag}, $\nabla f(\bm{w}^t)$ can be approximated by recent gradients or weight differences since $f(\bm{w}^t)$ is \textit{L}-smooth.
Then we define 
\begin{align} \label{eq:gradient_diff}
    \nabla f(\bm{w}^t) \approx  \frac{1}{\eta ^2}\sum _{k=1}^K \delta_k \norm{\bm{w}^{t+1-k}- \bm{w}^{t-k}}^2,
\end{align}
where $\{ \delta_k \}_{k=1}^K$ and $K$ are constants.
It is also noted that
\begin{align} \label{eq:relax_grad_diff}
    &\norm{ \sum _{i \in \mathcal{N}_e^t} \frac{D_i}{D} \left [ \nabla f_i(\widetilde{\bm{w}}_i^t) - \nabla f_i(\bm{w}^t)\right] }^2 \notag\\
    &\leq | \mathcal{N}_e^t | \sum_{i \in \mathcal{N}_e^t} \norm{ \nabla f_i(\widetilde{\bm{w}}_i^t) - \nabla f_i(\bm{w}^t) }^2.
\end{align}
With (\ref{eq:lemma1_new})-(\ref{eq:relax_grad_diff}),
the condition (\ref{eq:comm_cond}) can be easily formed to decide if uploading fresh models.
\end{remark}

Then we conclude the convergence rate of CEFL algorithm as follows.
\begin{theorem}[Convergence rate]\label{theorem1}
Let Assumptions \ref{ass:smoothness}-\ref{assump:strong_convex} hold and $L, \mu$ be defined therein, our proposed CEFL over wireless {\color{black}IoT} networks in Algorithm \ref{algorithm:cefl} can achieve a strong expected linear rate, i.e.,
\begin{equation} \label{eq:theorem}
    \begin{aligned}
        \mathbbm{E} \left[ f(\bm{w}^{t+1}) -f(\bm{w}^*) \right] \leq
        \left(1-\rho\right) \mathbbm{E}\left[ f(\bm{w}^t) -f(\bm{w}^*)\right],
    \end{aligned}
\end{equation}
where $\rho$ is a positive constant satisfying the following condition
\begin{equation} \label{eq:rho_condition}
    \rho \leq \frac{\mu}{L}\sum_{i\in \mathcal{N}_e^t}\frac{D_i \left (1- p_i^t \right)}{D},
\end{equation}
with $p_i^t$ denoting the outage probability during uploading.
\end{theorem}
\begin{IEEEproof}
The proof is detailed in Appendix \ref{secondAppendix}.
\end{IEEEproof}
\begin{remark}
Theorem \ref{theorem1} implies that conditioned on (\ref{eq:rho_condition}), the proposed CEFL still exhibits the same order of convergence rate as that of the original GD method even though some communications are skipped in CEFL. 
Theorem \ref{theorem1} also presents the impact of wireless factors on the convergence properties i.e., outage probability.
Thus, with $p_i^t \rightarrow 0$, the proposed CEFL algorithm will converge in a strong convergence rate. 
Otherwise, the linear convergence rate does not hold any longer.
Note that Theorem \ref{theorem1} provides a sufficient condition to guarantee the convergence speed of the proposed CEFL approach.
\end{remark}

\subsection{Communication Analysis}
Next, we analyze the communication cost based on the linear convergence rate.
In following analysis, communication of model parameters between the PS and client is taken as 1 unit of communication.
\begin{corollary}\label{colloary1_total_comm_round} 
Assume that positive constant $\rho$ meets (\ref{eq:rho_condition}).
To realize an expected convergence of $f(\bm{w})$ under an accuracy threshold $\epsilon$, i.e., $\mathbbm{E} \left[ f(\bm{w}^{t}) -f(\bm{w}^*) \right] \leq \epsilon $, the total number of communication rounds $T_{\textit{total}}$ for CEFL algorithm is lower bounded by
$    T_{\textit{total}}\geq \left \lceil \log_{1-\rho} \frac{\epsilon}{\mathbbm{E}\left[ f(\bm{w}^{1})\right]}\right \rceil $.
\begin{IEEEproof}
If we assume $\rho=\frac{\mu}{L}\sum_{i\in \mathcal{N}_e^t}\frac{D_i \left (1- p_i^t \right)}{D}$, we have
\begin{align}
    \mathbbm{E}\left[ f(\bm{w}^{t+1})-f(\bm{w}^*)\right] &\leq \left( 1- \rho \right)\mathbbm{E}\left[ f(\bm{w}^t)-f(\bm{w}^*)\right]\notag\\
    &\leq \left( 1- \rho \right)^2 \mathbbm{E}\left[ f(\bm{w}^{t-1})-f\bm{w}^*)\right] \notag\\
    &~...\notag\\
    &\leq \left( 1- \rho \right)^t \mathbbm{E}\left[ f(\bm{w}^{1})-f(\bm{w}^*)\right].
\end{align}
To achieve a pre-defined deviation defined by $f(\bm{w}^{t})-f(\bm{w}^*) \leq \epsilon$, it is sufficient to have
\begin{align}
    \left( 1- \rho \right)^t \mathbbm{E}\left[ f(\bm{w}^{1})-f(\bm{w}^*)\right] \leq \epsilon.
\end{align}
Then, we can further derive
\begin{align}
    t &\geq \log_{1-\rho} \frac{\epsilon}{\mathbbm{E}\left[ f(\bm{w}^{1})-f(\bm{w}^*)\right]}\notag\\
    &\geq \log_{1-\rho} \frac{\epsilon}{\mathbbm{E}\left[ f(\bm{w}^{1})\right]}.
\end{align}
Since the convergence round must be an integer, we get the result in Corollary \ref{colloary1_total_comm_round}.
\end{IEEEproof}
\end{corollary}

\begin{corollary}[Communication cost] \label{corollary2}
Let (\ref{eq:rho_condition}) holds, under the same conditions as Corollary \ref{colloary1_total_comm_round}, with deviation defined by $f(\bm{w}^{t})-f(\bm{w}^*) \leq \epsilon$,
the communication cost of the proposed CEFL algorithm is $O\left ( \log{\frac{1}{\epsilon}} \right)$.
\end{corollary}
\begin{IEEEproof}
With the formula of change of base of logarithms, Corollary \ref{corollary2} can be easily obtained.
\end{IEEEproof}
\begin{remark}
Compared with $N$ activated clients per communication round in \textit{FedAvg}-based FL algorithm, only a subset $\mathcal{N}_e^t$ of clients ($\left |\mathcal{N}_e^t \right| \leq N$) is active to upload fresh models in the CEFL-based approach. 
Corollary \ref{corollary2} shows that for the proposed CEFL algorithm, the total communication cost under an accuracy threshold $\epsilon$ is reduced to $O\left ( \log{\frac{1}{\epsilon}} \right)$.
\end{remark}
 
\section{Simulation Results and Analysis} \label{sec:results}
In this section, we evaluate the performance of the proposed CEFL approach under real datasets.
\subsection{Simulation Setup}
\begin{figure*}[t] 
 	\centering
 	\vskip -0.1in
	\subfloat[ ]{\hspace*{1mm}\includegraphics[width=60mm]{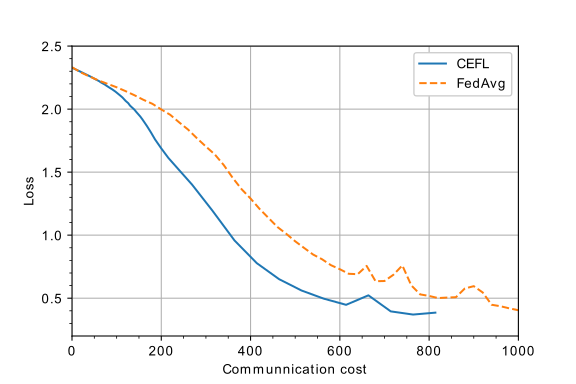}\label{fig_loss_vs_cost}}
	\subfloat[ ]{\hspace*{-2mm}\includegraphics[width=60mm]{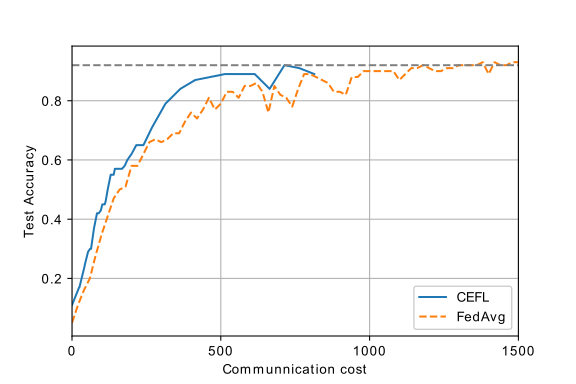}\label{fig_acc_vs_cost}}
 	\subfloat[ ]{\hspace*{-2mm}\includegraphics[width=60mm]{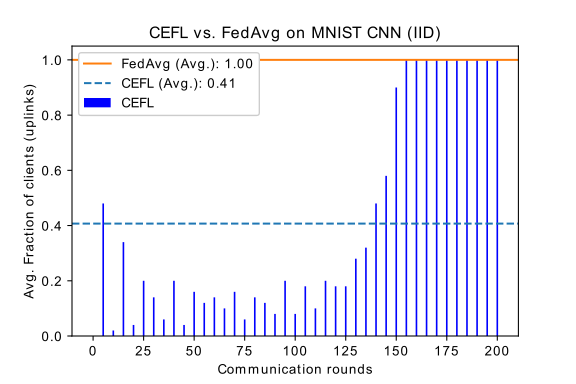}\label{fig_per_cost}} \\
 	\subfloat[ ]{\hspace*{1mm}\includegraphics[width=60mm]{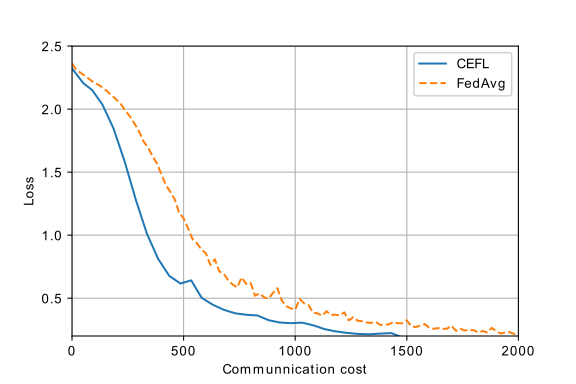}\label{fig_loss_vs_cost_v2}}
 	\subfloat[ ]{\hspace*{-2mm}\includegraphics[width=60mm]{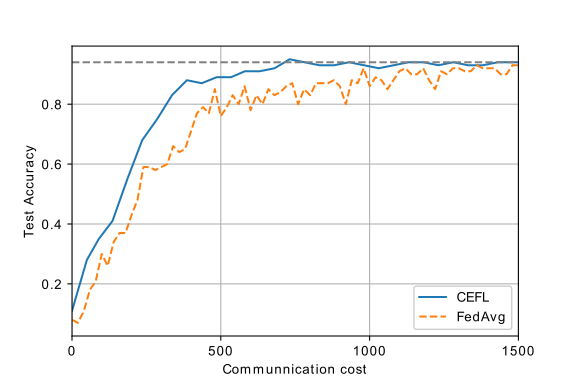}\label{fig_acc_cost_v2}}
 	\subfloat[ ]{\hspace*{-2mm}\includegraphics[width=60mm]{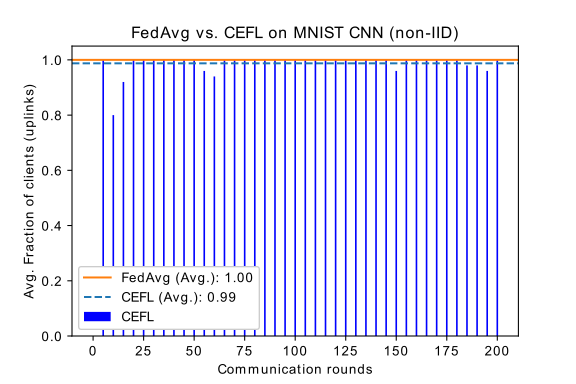}\label{fig_per_cost_v2}}
 	
    \caption{Training neural network for classification on MNIST dataset:  (a): communication overhead comparison (IID); (b) accuracy comparison (IID); (c) percentage of the involved clients (IID); (d) communication overhead comparison (non-IID); (e) accuracy comparison (non-IID); (f) percentage of the involved clients (non-IID).}
    \label{fig:learning_performance}
 \end{figure*}



For simulations, we consider a typical single-cell wireless {\color{black}IoT} network that consists of $N=10$ edge clients and a BS located at its center similar to the example network shown in Fig. \ref{fig_network}.
We assume that the BS has two ring-shaped boundary regions. 
The inner and outer boundaries have radii of 10 m and 500 m, respectively. 
The $N$ clients are uniformly and randomly distributed between the two boundaries, where the distance (in meter) between client $i$ and the BS is denoted as $d_i$. 
The wireless channels from each client to the BS follow i.i.d. Rayleigh fading with the total allowed bandwidth $B= \text{ 20 MHz}$, and the channel $h_i^t$ is modeled as
\begin{equation}
    h_i^t = \sqrt{ L(d_i)} o_i^t,
\end{equation}
where $o_i^t \sim \mathcal{CN}\left(0, \sigma ^2\right)$ is the small-scale fading coefficient of the link between client $i$ and BS, and $L(d_i)=\beta_0 (d_i)^{-\alpha}$ is the distance-dependent pathloss with exponent $\alpha$ and coefficient $\beta_0$ \cite{9187874}. 
$\beta_0$ is a frequency-dependent constant, which is set as $(\frac{c}{4\pi f_c})^2$ with $c=3\times10^8 \text{ m/s}$ and the carrier frequency $f_c=3 \text{ GHz}$.
Then, the $p_i^t$ can be formulated as
\begin{equation}
    p_i^t=Pr(\tau_i > \Gamma^t)=
    1 -\exp{\left(-\frac{Q_i^t}{P_i^t}\right)},
\end{equation}
where $Q_i^t= \frac{B_i^t N_0}{L(d_i) \sigma ^2}\left( 2^{\frac{S}{B_i^t \Gamma^t}} -1\right)$.
Unless specifically stated otherwise, other parameters are given in Table \ref{tab:simu_para}, following the studies in \cite{chen_fl1, wang2021federated, joint_schedul}.
To investigate the performance especially in communication efficiency, we compare our CEFL approach against the vanilla FL approach \cite{fedavg}, over wireless networks, among which $\textit{FedAvg}$ method is adopted.
{\color{black}
For the target model, we consider a convolutional neural network (CNN) architecture which has two $5\times5$ convolutional layers (the first with 10 channels, the second with 20 channels, each followed by ReLU function), each followed by a $2\times2$ max pooling layer, a fully connected layer with 500 units and ReLU function, and a final softmax output layer.
}
Our model was simulated by Tensorflow in Python 3.7 and all experiments were carried out on the environment with the following hardware specifications: CPU Intel Core i5 @2.3 GHz; RAM 16 GB.

\begin{table}[!t]
    \caption{Simulation Parameters \cite{simulation_para}}
    \label{tab:simu_para}
    \centering
    \fontsize{9}{8}\selectfont
    \begin{tabular}{|c|c|c|c|}
        \hline 
        \textbf{Parameter} & \textbf{Value}&\textbf{Parameter} & \textbf{Value} \\
        \hline 
        $f_c$ & $3\text{ GHz}$ & $B$ & $20 \text{ MHz}$ \\ \hline
        $\alpha$  & $2.9$ & $P^{\text{max}}$ & $20 \text{ dBm}$\\ \hline
        $N_0$ & $-174 \text{ dBm/Hz}$ & $P^{\text{min}}$ & $0 \text{ dBm}$\\ 
        \hline
    \end{tabular}
\end{table}

\subsection{Simulation Results}
We evaluate the performance of the proposed approach via the MNIST dataset for handwritten digits classification \cite{mnist_data}. 
The MNIST dataset has 60, 000 training images and 10, 000 testing images of the 10 digits.
We adopt the common assumption that each client is connected with the equal amount of training data samples and the local training samples are non-overlapping with each other [15] [20].
Besides, different data distributions of training samples are considered, both i.i.d. case and non-i.i.d. case.
For the i.i.d. case, the original dataset is first uniformly partitioned into $N$ pieces and each client is assigned one piece. 
While for the non-i.i.d. case, the original training dataset is first partitioned into $N$ pieces according to the label order, and each piece is then randomly partitioned into 2 shards (i.e., $2N$ shards in total).
Finally, each of $N$ clients is assigned 2 shards with different label distribution.

 \begin{figure}[t] 
 	\centering
 	\vskip -0.1in
	\subfloat[ ]{\hspace*{1mm}\includegraphics[width=88mm]{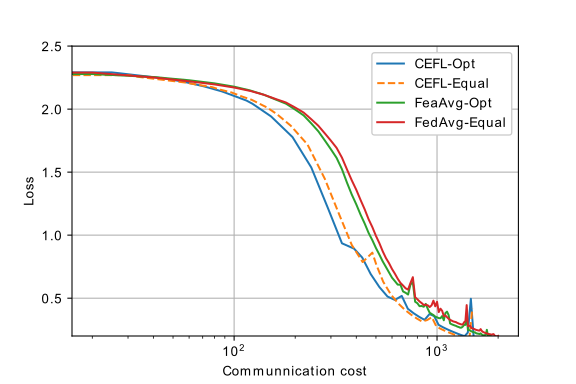}\label{fig_ham1}}\\
	\subfloat[ ]{\hspace*{-2mm}\includegraphics[width=88mm]{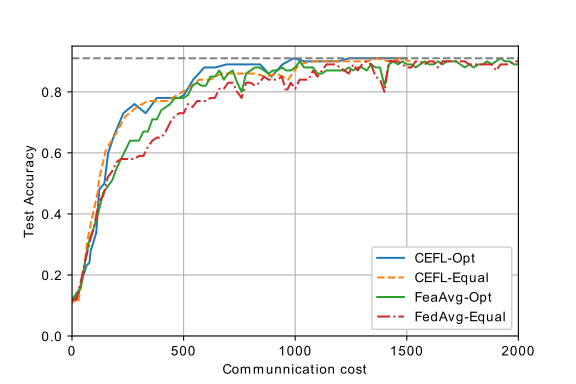}\label{fig_b1}}
    \caption{Impact of resource allocation scheme on the convergence of the proposed CEFL on MNIST dataset: (a) training loss comparison;  (b) test accuracy comparison.}
    \label{fig:rsa_impact}
 \end{figure}

The performance of the proposed CEFL approach for classification on MNIST dataset is evaluated in Fig. \ref{fig:learning_performance} by training a CNN model over cumulative communication overhead.
We first report the results for i.i.d. data partition case.
The corresponding experimental results about the training loss, test accuracy, and the utilization of clients are shown in Fig. 2(a), Fig. 2(b), and Fig. 2(c), respectively.
It is observed that under the same total communication overhead, the proposed CEFL approach performs better than the vanilla FL method (\textit{FedAvg}).
This is because that less informative messages (i.e., local FL models) from clients are restricted to upload to the PS in CEFL approach whilst all messages are transmitted to the PS for updating in \textit{FedAvg} method at each communication round.
We use an intuitive explanation as shown in Fig. \ref{fig:learning_performance}(c) to showcase the effectiveness of CEFL on selectively uploaded local models. 
In Fig. \ref{fig:learning_performance}(c), one blue stick refers to the percentage of participated clients to upload local models at each communication round.
In the initial several communication rounds, communication events happen sparsely. While during the late communication rounds (150-th to 200-th), almost all messages (i.e., local FL models) from clients are critical and selected to transmit in our proposed policy, implying that client scheduling policy works better at the first half communication rounds.
Similar performance results have also been observed in Fig. \ref{fig:learning_performance}(d)-Fig. \ref{fig:learning_performance}(f), where non-i.i.d. case on the MNIST training dataset is taken into account. 
It is worth noting that compared with that of i.i.d. case, the percentage of participated clients is much higher in non-i.i.d. case.

Furthermore, we evaluate the impact of the resource allocation scheme on the learning performance in Fig. \ref{fig:rsa_impact}.
For performance comparison, we also implement the equal resource allocation approach of \cite{chen2020convergence} as a benchmark.
For this allocation scheme, in each communication round $t$, both the transmission power and its bandwidth of each client are identical, i.e., $B_i^t = B/N, P_i^t = P^{\text{max}}, \forall i \in \mathcal{N}$.
In Fig. \ref{fig:rsa_impact}(a), we first show how resource allocation scheme affects the convergence behavior of the global FL model training in terms of the value of the training loss.
As the communication cost increases, the training losses of the considered algorithms decrease at different rates, whilst the proposed CEFL framework consisting of the new client scheduling policy and the LS based resource allocation (denoted as 'CEFL-Opt') achieves the lowest loss.
Fig. \ref{fig:rsa_impact}(b) also represents that the proposed LS based CEFL algorithm achieves the highest test accuracy among all schemes under the same communication budget on MNIST.
This is reasonable since both client selection and resource allocation are taken into account in the proposed CEFL framework so as to reduce the effect of wireless transmission errors in FL.


\section{Conclusions} \label{conclusion}
We have studied the joint optimization problem of communication and resources with federated learning over wireless {\color{black}IoT} networks, in which both client selection and resource allocation are considered. 
A CEFL framework was proposed combined with a new client scheduling policy and an LS based allocation method, respectively. 
We showed that the presented LS approach was able to provide an optimal solution for bandwidth and power allocation and the convergence and communication properties of the proposed CEFL algorithm were also theoretically analyzed.
Extensive experimental results revealed that the proposed CEFL algorithm outperforms the state-of-the-art baseline method both in communication overheads and learning performance under different data distributions.
Besides, the proposed CEFL framework can effectively schedule clients according to both the learned model parameter characteristics and wireless channel dynamics.

\appendices
\section{Supporting Lemma \ref{lemma_3}} \label{firstAppendix}
Before proving Theorem 1, we first introduce the following Lemma \ref{lemma_3}.
\begin{lemma}\label{lemma_3}
   Suppose that the iterates $\{\bm{w}^{t }\}$ of problem (1) are generated by full gradient descent over wireless {\color{black}IoT} networks: $\bm{w}^{t+1}=\bm{w}^t - \eta \bm{g}^t $ with $\bm{g}^t \overset{\Delta}{=} \nabla f(\bm{w}^t) + \bm{e}^t$ and learning rate $\eta^=\frac{1}{L}$ and the error $\bm{e}^t$ meets 
   \begin{align} \label{eq:error_condi}
       0\leq  \mathbbm{E} \left [ \norm{ \bm{e}^t}^2\right ] \leq 2L(\frac{\mu}{L}-\rho)\mathbbm{E}\left[ f(\bm{w}^t)-f(\bm{w}^*)\right],
   \end{align}
where $\rho \leq \frac{\mu}{L}$ is a positive constant.
Under Assumptions \ref{ass:smoothness}-\ref{assump:strong_convex}, the following inequality holds  
{\color{black}
\begin{align}\label{eq:lemma3}
        \mathbbm{E} \left[ f(\bm{w}^{t+1}) -f(\bm{w}^*) \right] \leq
        \left(1-\rho\right) \mathbbm{E}\left[ f(\bm{w}^t) -f(\bm{w}^*)\right].
 \end{align} 
}
\end{lemma}
\begin{IEEEproof}
Using Assumption \ref{ass:smoothness}, we can derive
\begin{align} \label{eq:derived_from_smooth}
    &f(\bm{w}^{t+1})\notag\\
    &\leq f(\bm{w}^t) + \langle \bm{w}^{t+1}-\bm{w}^t,  \nabla f(\bm{w}^t) \rangle +\frac{L}{2} \norm{\bm{w}^{t+1}-\bm{w}^t}^2 \notag\\
    &=f(\bm{w}^t) - \langle \eta \left( \nabla f(\bm{w}^t) + \bm{e}^t\right),  \nabla f(\bm{w}^t) \rangle +\frac{L}{2} \norm{\bm{w}^{t+1}-\bm{w}^t}^2 \notag\\
    &=f(\bm{w}^t) - \langle \frac{1}{L} \left( \nabla f(\bm{w}^t) + \bm{e}^t\right),  \nabla f(\bm{w}^t) \rangle \notag\\
    &\quad +\frac{L}{2} \norm{\frac{1}{L}\left( \nabla f(\bm{w}^t) + \bm{e}^t\right)}^2 \notag\\
    &=f(\bm{w}^t) - \frac{1}{2L} \norm{\nabla f(\bm{w}^t)}^2 +\frac{1}{2L} \norm{\bm{e}^t}^2.
\end{align}
Subtracting $f(\bm{w}^*)$ from both sides of (\ref{eq:derived_from_smooth}), it gives
\begin{align} \label{eq:derived_from_strong_convex}
   &f(\bm{w}^{t+1})-f(\bm{w}^*)\notag\\
   &\leq f(\bm{w}^t)-f(\bm{w}^*)- \frac{1}{2L} \norm{\nabla f(\bm{w}^t)}^2 +\frac{1}{2L} \norm{\bm{e}^t}^2\notag\\
   &\overset{(a)}{\leq}f(\bm{w}^t)-f(\bm{w}^*)-\frac{2\mu}{2L}\left[ f(\bm{w}^t)-f(\bm{w}^*)  \right]+\frac{1}{2L} \norm{e^t}^2\notag\\
   &=\left( 1-\frac{\mu}{L}\right)\left[ f(\bm{w}^t)-f(\bm{w}^*)  \right]+\frac{1}{2L} \norm{\bm{e}^t}^2,
\end{align}
where (a) holds due to (\ref{eq:ass_grad_inequality}).
Taking expectations of (\ref{eq:derived_from_strong_convex}), we can derive
\begin{align}
    &\mathbbm{E} \left[(\bm{w}^{t+1})-f(\bm{w}^*)\right] \notag\\
    &\leq \left( 1-\frac{\mu}{L}\right)\mathbbm{E}\left[ f(\bm{w}^t)-f(\bm{w}^*)  \right]+\frac{1}{2L} \mathbbm{E}\left[\norm{e^t}^2\right] \notag\\
    &\overset{(b)}{\leq}\left( 1-\frac{\mu}{L} +\frac{\mu}{L} - \rho \right) \mathbbm{E}\left[ f(\bm{w}^t)-f(\bm{w}^*) \right]\notag\\
    &=\left( 1- \rho \right)\mathbbm{E}\left[ f(\bm{w}^t)-f(\bm{w}^*) \right],
\end{align}
where (b) uses the condition (\ref{eq:error_condi}).
This completes the proof of Lemma \ref{lemma_3}.

\end{IEEEproof}

\section{Proof of Theorem \ref{theorem1}}
\label{secondAppendix}

\begin{IEEEproof}
Based on the result given in Lemma \ref{lemma_3}, we present the proof for Theorem 1 in detail.
Combing (\ref{eq:gd_method}) with (\ref{eq:cefl_sum}), we have
\begin{align}
    \bm{w}^{t+1} - \bm{w}^{t} = -\eta \left( \nabla f(\bm{w}^t) + \bm{e}^t \right),
\end{align}
where $\bm{e}^t$ is gradient deviation caused by the PS at communication round $t$ that it uses old copy of local FL model from client $i \in{\mathcal{N}}$ when the newly local FL model can not be successfully received. 
Let $\mathcal{N}_e^t$ and $\mathcal{N}_c^t$ be the sets of clients that \textit{do} and \textit{do not} communicate with the PS, respectively.
In particular, $\bm{e}^t$ can be expressed as (\ref{eq:e_def}), shown on the upper of the next page.
\begin{figure*}[t] 
\begin{align} \label{eq:e_def}
    e^t &= - \nabla f(\bm{w}^t) + \frac{\sum_{i\in \mathcal{N}_e^t}D_i a_i^t\nabla f_i(\bm{w}^t) +\sum_{i\in \mathcal{N}_e^t} D_i \left( 1-a_i^t\right) \nabla f_i(\widetilde{\bm{w}}_i^t) + \sum_{i\in \mathcal{N}_c^t} D_i \nabla f_i(\widetilde{\bm{w}}_i^t) }{D}\notag\\
    &\overset{(c)}{=}\frac{-\sum_{i\in \mathcal{N}_e^t}D_i \left( 1-a_i^t\right) \nabla f_i(\bm{w}^t) +\sum_{i\in \mathcal{N}_e^t} D_i \left( 1-a_i^t\right) \nabla f_i(\widetilde{\bm{w}}_i^t) +\sum_{i\in \mathcal{N}_c^t} D_i \nabla f_i(\widetilde{\bm{w}}_i^t)  - \sum_{i\in \mathcal{N}_c^t} D_i \nabla f_i(\bm{w}^t)}{D}\notag\\
    &= \sum_{i\in \mathcal{N}_e^t}\frac{D_i \left( 1-a_i^t\right)}{D} \left( \nabla f_i(\widetilde{\bm{w}}_i^t) -\nabla f_i(\bm{w}^t)\right) +\sum_{i\in \mathcal{N}_c^t} \frac{D_i}{D} \left(\nabla f_i(\widetilde{\bm{w}}_i^t) -\nabla f_i(\bm{w}^t)\right),
\end{align}
where (c) holds because of $\nabla f(\bm{w}^t)=\left( \sum_{i\in \mathcal{N}_e^t} D_i \nabla f_i(\bm{w}^t) + \sum_{i\in \mathcal{N}_c^t} D_i \nabla f_i(\bm{w}^t) \right) / D$.

\hrulefill
\end{figure*}
Thus, with $\eta^t=\frac{1}{L}$, we have
\begin{align} \label{eq:weight_diff}
    &\bm{w}^{t+1} - \bm{w}^{t} \notag\\
    &= -\frac{1}{L} \left( \nabla f(\bm{w}^t) + \sum_{i\in \mathcal{N}_e^t}\frac{D_i \left( 1-a_i^t\right)}{D} \left( \nabla f_i(\widetilde{\bm{w}}_i^t) -\nabla f_i(\bm{w}^t)\right) \right . \notag\\
    &\qquad\qquad \left . +\sum_{i\in \mathcal{N}_c^t} \frac{D_i}{D} \left(\nabla f_i(\widetilde{\bm{w}}_i^t) -\nabla f_i(\bm{w}^t)\right) \right).
\end{align}
Then, by taking expectations and norms in both sides of (\ref{eq:e_def}), we have
\begin{align} \label{eq:e_expectation}
    &\mathbbm{E} \left[\norm{\bm{e}^t}^2 \right] \notag\\
    &= \mathbbm{E} \left[\left \lVert
    \sum_{i\in \mathcal{N}_e^t}\frac{D_i \left( 1-a_i^t\right)}{D} \left(
    \nabla f_i(\reallywidehat{\bm{w}}_i^{t}) -\nabla f_i(\bm{w}^t)\right) \right. \right . \notag\\
    &\quad\qquad\left .  \left .  +\sum_{i\in \mathcal{N}_c^t} \frac{D_i}{D} \left(\nabla f_i(\reallywidehat{\bm{w}}_i^{t}) -\nabla f_i(\bm{w}^t)\right) \right \rVert ^2 \right ] \notag\\
    &\leq\mathbbm{E} \left[\left \lVert
    \sum_{i\in \mathcal{N}_e^t}\frac{D_i \left( 1-a_i^t\right)}{D} \norm{
    \nabla f_i(\reallywidehat{\bm{w}}_i^{t}) -\nabla f_i(\bm{w}^t)} \right. \right.  \notag\\
    &\qquad\quad \left. \left .+\sum_{i\in \mathcal{N}_c^t} \frac{D_i}{D} \norm{\nabla f_i(\reallywidehat{\bm{w}}_i^{t}) -\nabla f_i(\bm{w}^t)} \right \rVert ^2 \right] \notag\\
    &\overset{(e)}{\leq}\mathbbm{E} \left[\norm{
    \sum_{i\in \mathcal{N}_e^t}\frac{D_i \left( 1-a_i^t\right)L G_{\mathcal{X}}}{D}  +\sum_{i\in \mathcal{N}_c^t} \frac{D_i L G_{\mathcal{X}}}{D} }^2 \right] \notag\\
    &=\mathbbm{E} \left[\norm{   L  G_{\mathcal{X}} \left(  \sum_{i\in \mathcal{N}}\frac{D_i}{D} -
    \sum_{i\in \mathcal{N}_e^t} \frac{D_i a_i^t}{D}
    \right)   }^2 \right] \notag\\
    &\overset{(f)}{\leq} L^2 G_{\mathcal{X}}^2 \left(  1- \sum_{i\in \mathcal{N}_e^t} \frac{D_i \mathbbm{E} \left[  a_i^t\right]}{D} \right) \notag\\
    & \overset{(g)}{=} L^2 G_{\mathcal{X}}^2\left(  1- \sum_{i\in \mathcal{N}_e^t} \frac{D_i \left (1- p_i^t \right) }{D} \right),
\end{align}
where (e) is due to Assumption 1, (f) is because of $\sum_{i\in \mathcal{N}}D_i = D$ and (g) is based on $\mathbbm{E} \left[  a_i^t\right] = 1 - p_i^t$, respectively.
According to Lemma \ref{lemma_3}, if let (\ref{eq:error_condi}) always holds, we have
\begin{align}
    \mathbbm{E} \left[\norm{\bm{e}^t}^2 \right]&\leq L^2 G_{\mathcal{X}}^2\left(  1- \sum_{i\in \mathcal{N}_e^t} \frac{D_i \left (1- p_i^t \right)}{D} \right) \notag\\
    &\leq 2L(\frac{\mu}{L}-\rho)\mathbbm{E}\left[ f(\bm{w}^t)-f(\bm{w}^*)\right].
\end{align}
That is, to guarantee (\ref{eq:lemma3}), the available $\rho$ must satisfy
\begin{align} \label{eq:rho_cal}
    \rho &\leq \frac{\mu}{L} - \frac{L G_{\mathcal{X}}^2  \left(  1- \sum_{i\in \mathcal{N}_e^t} \frac{D_i \left (1- p_i^t \right)}{D} \right)}{2\mathbbm{E}\left[ f(\bm{w}^t)-f(\bm{w}^*)\right]}.
\end{align}
According to (\ref{ass:smooth_eq1}) in Assumption \ref{ass:smoothness} and (\ref{eq:ass_grad_inequality}) in Assumption \ref{assump:strong_convex}, we have
\begin{align}
    \mathbbm{E}\left[ f(\bm{w}^t)-f(\bm{w}^*)\right]\leq \frac{\norm{\nabla f(\bm{w}^t)}^2}{2\mu}\leq
    \frac{L^2 G_{\mathcal{X}}^2}{2\mu}.
\end{align}
Then, (\ref{eq:rho_cal}) can be derived as
\begin{align}
    \rho &\leq \frac{\mu}{L} - \frac{\mu \left( 1- \sum_{i\in \mathcal{N}_e^t} \frac{D_i \left (1- p_i^t \right)}{D} \right)}{L} \notag\\
    &=\frac{\mu}{L}\sum_{i\in \mathcal{N}_e^t}\frac{D_i \left (1- p_i^t \right)}{D}.
\end{align}
This completes the proof of Theorem \ref{theorem1}.

\end{IEEEproof}

\bibliography{reflibb}
\bibliographystyle{IEEEtran} 


\end{document}